%% file: main.tex
\documentclass[10pt]{article}
\usepackage[utf8]{inputenc}
\usepackage[T1]{fontenc}
\usepackage[english]{babel}
\usepackage{faider_preprint}
\usepackage{lmodern,microtype,csquotes}
\usepackage{amsmath,amssymb,amsthm,mathtools}
\usepackage{graphicx,subcaption,booktabs,multirow,adjustbox}
\usepackage{algorithm,algorithmic}
\usepackage{xcolor}
\usepackage[colorlinks=true,citecolor=blue,linkcolor=blue,urlcolor=blue]{hyperref}
\usepackage[capitalize,nameinlink]{cleveref}
\usepackage[style=numeric,sorting=none,backend=biber,maxbibnames=6]{biblatex}
\AtBeginBibliography{\sloppy}
\newcommand{\faider}{\textsc{FAIDER}}
\newcommand{\Emb}{\operatorname{Emb}}
\newcommand{\Ncal}{\mathcal{N}}
\newcommand{\Lcal}{\mathcal{L}}
\newcommand{\Exp}{\mathbb{E}}
\newcommand{\one}{\mathbf{1}}
\newcommand{\err}{\mathrm{err}}
\newcommand{\Xcal}{\mathcal{X}}
\newcommand{\Acal}{\mathcal{A}}
\newcommand{\Ycal}{\mathcal{Y}}

\DeclareMathOperator*{\argmax}{arg\,max}

\theoremstyle{definition}
\newtheorem{definition}{Definition}
\newtheorem{assumption}{Assumption}
\newtheorem{lemma}{Lemma}
\newtheorem{prop}{Proposition}

\title{FAIDER: Fairness-Aware Item-Level Decoding for Large Language Model Recommenders}
\author{Ziwei Wu \qquad Jingrui He\\University of Illinois Urbana-Champaign\\\texttt{ziweiwu2@illinois.edu} \qquad \texttt{jingrui@illinois.edu}}
\date{}
\begin{document}
\maketitle
\begin{abstract}
Large language model (LLM) recommenders can produce semantically different recommendations when a user's protected attribute changes, even if the interaction history remains fixed. Controlling this disparity is challenging when model weights and full-vocabulary logits are inaccessible. We propose \faider{} (Fairness-Aware Item-level DEcoding for Recommendation), which incorporates fairness into list construction by selecting complete item candidates rather than manipulating token-level logits or repeatedly repairing finished outputs. The central idea is to separate the operational criterion for a violation from how strongly generation is constrained. Conformal-inspired empirical calibration fixes a cross-group semantic-disparity threshold, while online decoding applies either hard filtering or a soft penalty adjusted through dual-inspired feedback to target a prescribed violation rate. Separately, we establish a pathwise guarantee: under a uniform enforceability assumption, the soft decoder's empirical violation rate exceeds its target by at most $O(1/T)$. Experiments across recommendation benchmarks and LLM backbones show that Soft \faider{} improves both measured semantic fairness and ranking quality over the evaluated prompt-repair baselines. It also generally achieves a more favorable fairness--utility balance than the hard variant, indicating that stricter per-recommendation constraints do not necessarily yield lower measured disparity.
\end{abstract}
\input{c7_intro}
\input{c7_related}
\input{c7_method}
\input{c7_analysis}
\input{c7_exp}
\input{c7_conclusion}
\printbibliography
\end{document}

%% file: c7_intro.tex
\section{Introduction}
Large language models (LLMs) make it possible to express recommendation as a
language-based decision process. Text-to-text frameworks such as P5 unify
recommendation tasks within a generative interface \cite{geng2022p5}, while
zero-shot rankers show that pretrained LLMs can rank candidates from interaction
histories without recommendation-specific training \cite{hou2024large}. The
broader generative recommendation literature uses this interface to connect
behavioral context, item semantics, and natural-language requests
\cite{li2024generative}. Yet sensitivity to contextual language also creates a
fairness concern: demographic information in a user profile can affect
recommendation behavior even when the interaction history is held fixed
\cite{zhang2023chatgpt,ma2026rethinking}.

Fairness in recommendation encompasses distinct objectives, including provider
exposure, consumer utility, and individual-level consistency
\cite{singh2018exposure,vassoy2024consumer,deldjoo2024landscape}. Language-based
systems introduce additional targets, such as fairness in personalized
explanations \cite{wang2023coffee}. We focus on \emph{minimal-attribute-change
semantic stability}: changing a protected attribute while holding a user's
interaction history fixed should not substantially alter the semantic content
of the recommended items. This operational criterion concerns sensitivity to
an input perturbation; it does not identify a structural causal counterfactual
in the sense of \textcite{kusner2017counterfactual}. Nor should similarity
between recommendation lists be equated with equal user benefit, a distinction
highlighted by consumer-fairness evaluation \cite{deldjoo2025cfairllm}. We
therefore assess semantic disparity alongside recommendation quality.

The available interventions depend on model access. Tuning-based frameworks
adapt pretrained models to recommendation data \cite{bao2023tallrec}, and UP5
learns continuous prompts to address recommendation fairness
\cite{hua2024up5}. Such approaches require a trainable component. We instead
study a fixed recommender that exposes sampled item continuations and their
log-probabilities, but not model weights or full-vocabulary logits. The design
problem is to control recommendation behavior through the item proposals and
scores available at inference time.

One response is to revise prompts after inspecting generated outputs. FACTER
combines conformal thresholding with prompt updates informed by previous
fairness violations \cite{fayyazi2025facter}. More broadly, iterative
self-refinement uses feedback on an initial response to guide subsequent
revisions without additional model training \cite{madaan2023selfrefine}.
These approaches motivate inference-time intervention, but their feedback
operates on outputs that have already been generated. We ask a complementary
question: \emph{can a fairness criterion guide which items are committed to a
recommendation list as the list is being constructed?}

Decoding-time control offers a relevant starting point. \cite{Dathathri2020Plug,yang2021fudge,liu2021dexperts} guide generation through hidden-state updates or adjustments to token distributions.
Their control rules, however, are not directly available through an interface
restricted to sampled item continuations and their likelihoods. Moreover, the
semantic disparity of a recommendation list is evaluated over items, rather
than isolated tokens. This motivates an item-level decision rule: evaluate
each proposed item's effect on the partial list, select a continuation, and
condition subsequent generation on that commitment.

We propose \faider{} (\textbf{F}airness-\textbf{A}ware \textbf{I}tem-Level \textbf{DE}coding for \textbf{R}ecommendation),
built around a separation between \emph{the criterion for a violation} and
\emph{the strength of its enforcement}. Offline calibration fixes an empirical
threshold on embedding-based cross-group disparity, a decoding surrogate for the paired attribute-change metric.
During generation, the hard decoder
filters candidates against this threshold and abstains if no feasible beam
remains. The soft decoder instead balances candidate likelihood against
disparity through a penalty whose weight is updated from observed violations.
A violation thus changes how strongly the decoder intervenes, rather than
changing the cutoff that determines whether a violation occurred. Keeping
these roles separate provides a stable operational criterion while allowing
the fairness--utility trade-off to adapt.

This separation also exposes a statistical issue. Split conformal prediction
gives calibrated thresholds a coverage interpretation when calibration and
test scores are exchangeable
\cite{angelopoulos2023gentle}; related ideas extend to risk control and
set-valued language generation
\cite{angelopoulos2024risk,quach2024conformal}. Our procedure uses the same
calibration set to supply reference outputs and threshold-selection scores,
while keeping the test set separate. This reference reuse does not
automatically preserve the score exchangeability required by the standard
argument; we therefore use the quantile as a conformal-inspired empirical
cutoff, without claiming finite-sample test coverage.
Adaptive decoding introduces an additional change to the output-selection
mechanism.
Motivated by feedback-based approaches to adaptive conformal inference and
online risk control \cite{gibbs2021adaptive,feldman2023risk}, we analyze
realized violations separately from offline calibration. Under an explicit
uniform enforceability assumption on candidate pools and the selection rule,
we establish a pathwise $O(1/T)$ upper bound on the soft decoder's excess
empirical violation rate for arbitrary query sequences. The guarantee is
conditional: selection alone cannot ensure compliance when acceptable
continuations are unavailable.

Experiments across recommendation benchmarks and LLM backbones show that
Soft \faider{} improves measured semantic fairness and ranking quality over
the evaluated prompt-repair baselines. Its generally more favorable fairness--utility balance than the hard variant also shows that stricter per-recommendation constraints do not necessarily yield lower measured
disparity. Our contributions are:
\begin{itemize}
  \interlinepenalty=10000
  \item Item-level generation control. We introduce a fairness-aware
  decoder for fixed LLM recommenders using item proposals and candidate
  likelihoods, without full-vocabulary logits, LLM retraining, or iterative
  prompt repair.
  \item Separated calibration and enforcement. We combine a fixed
  empirical disparity threshold with hard and soft decoders, and establish
  a conditional pathwise violation-rate bound for the soft decoder through
  dual-inspired online feedback.
  \item Empirical characterization of the trade-off. We compare
  fairness and ranking quality across recommendation benchmarks and LLM
  backbones, examining when soft penalties offer a more favorable balance
  than strict filtering.
\end{itemize}

%% file: c7_related.tex
\section{Related Work}
\label{sec:related}

\paragraph{Fairness objectives and evaluation.}
Surveys organize fair recommendation by the stakeholders protected, the
quantities compared, and the stage at which intervention occurs
\cite{deldjoo2024landscape,vassoy2024consumer,jin2023survey,ma2026rethinking}.
Exposure-based approaches address the allocation of attention to ranked items
\cite{singh2018exposure}, whereas consumer-fairness evaluation considers
whether recommendations serve users' preferences equitably
\cite{deldjoo2025cfairllm}. These objectives are not interchangeable with
similarity between generated lists. Likewise, counterfactual fairness is
defined relative to a causal model \cite{kusner2017counterfactual}, rather
than an attribute edit alone. Our focus is narrower: semantic stability under
protected-attribute changes, assessed together with ranking quality. This
criterion complements, rather than replaces, exposure-based, preference-aware,
and causal notions of fairness.

\paragraph{Fairness interventions for LLM recommendation.}
UP5 learns continuous prefix prompts through adversarial training to suppress
sensitive information in user representations \cite{hua2024up5}. COFFEE
instead targets measure-specific counterfactual fairness in personalized
explanation generation \cite{wang2023coffee}, addressing the quality of
generated text rather than the semantic composition of item lists. FACTER is
closer to our setting: it combines adaptive conformal thresholding with prompts
updated from observed fairness violations, without retraining the recommender
\cite{fayyazi2025facter}. Iterative output refinement also appears in
general-purpose methods such as Self-Refine \cite{madaan2023selfrefine},
although its objective is not recommendation fairness. \faider{} uses a
different intervention point. It neither learns prefix prompts nor revises
completed lists through corrective prompting. Its soft variant uses observed
list-level violations to adjust the penalty applied during item selection,
while the calibrated disparity cutoff remains fixed.

\paragraph{Controlled generation and the decision unit.}
PPLM steers generation by applying attribute gradients to hidden states
\cite{Dathathri2020Plug}, while FUDGE uses predictions of future attributes
to modify next-token probabilities \cite{yang2021fudge}. DExperts combines
expert and anti-expert language-model scores \cite{liu2021dexperts}, and
GeDi derives token-level guidance from class-conditional language models
\cite{krause2021gedi}. These approaches demonstrate that a fixed base model
can be steered at inference time, but their control rules operate through
internal states or token distributions. \faider{} shares the idea of
generation-time guidance while changing its unit of action: it scores
complete item proposals through their effect on the partial recommendation
list, commits one, and conditions subsequent proposals on that choice. This
differs from re-ranking a completed list and requires candidate likelihoods,
not an accessible full-vocabulary distribution or modifiable hidden states.

\paragraph{Conformal calibration for generated outputs.}
Classical conformal prediction provides marginal coverage under
exchangeability and appropriate score construction
\cite{vovk2005algorithmic,angelopoulos2023gentle}. Conformal risk control
extends this principle to expected bounded monotone losses
\cite{angelopoulos2024risk}. In language generation, Conformal Language
Modeling calibrates sampling and filtering procedures to construct sets
containing an acceptable response with a prescribed probability
\cite{quach2024conformal}. Such set-level guarantees differ from controlling
the disparity of a single emitted recommendation list. Weighted conformal
methods can accommodate covariate shift when the required likelihood-ratio
information is available \cite{tibshirani2019covariate}, but do not by
themselves justify arbitrary changes to output selection. In \faider{},
the same calibration examples supply both reference outputs and scores for
threshold selection. We consequently use the quantile as a conformal-inspired
empirical cutoff, not a finite-sample coverage guarantee. Adaptive selection
of a single output is an additional reason that classical coverage results
cannot simply be transferred to the decoder.

\paragraph{Online feedback and violation control.}
Online methods complement marginal coverage with retrospective guarantees
based on observed outcomes. Adaptive conformal inference adjusts calibration
in response to coverage errors \cite{gibbs2021adaptive}, while online risk
control extends feedback-based regulation to broader losses
\cite{feldman2023risk}. Work on decaying step sizes further examines the
relationship between long-run coverage and pointwise behavior
\cite{angelopoulos2024online}. Soft \faider{} uses a related feedback
structure but updates a dual-inspired penalty weight while keeping the disparity
threshold fixed. Its guarantee concerns the realized frequency of threshold
violations and requires uniform enforceability of the candidate-selection
rule. Neither offline calibration nor the existence of a feasible recommendation
alone establishes that assumption. The resulting claim is conditional online
control of recommendation generation, rather than an unconditional extension
of conformal coverage to adaptively selected lists.

%% file: c7_method.tex
\section{Problem Setup}
Let $x\in\Xcal$ denote a user context, such as an interaction history,
$a\in\Acal$ a protected attribute, and $y\in\Ycal$ a recommendation
represented as a ranked list of item names. A fixed LLM recommender defines
$p_\theta(y\mid x,a)$. We assume access to sampled item continuations and
their log-probabilities, but not to model weights or full-vocabulary logits.
The selection rules below require candidate log-probabilities. A sample-only interface would require a different scoring rule.

We write $\Emb(\cdot)$ for a fixed text-embedding map into $\mathbb{R}^m$,
applied to the textual representations of user contexts and item lists.
Context embeddings are compared with other contexts, and list embeddings
with other lists.

\begin{definition}[Minimal-attribute-change fairness]
\label{def:mac}
A recommender satisfies minimal-attribute-change fairness at tolerance
$\varepsilon_{\mathrm{fair}}$ if, for protected-attribute changes $a\to a'$
with context fixed,
$\lVert \Emb(y(x,a)) - \Emb(y(x,a')) \rVert_2 \le \varepsilon_{\mathrm{fair}}$.
\end{definition}
This criterion expresses the evaluation objective: \emph{semantic stability}
when a protected attribute changes but the user context remains fixed.
In the experiments, CFR~\cite{fayyazi2025facter} aggregates the corresponding embedding
distances (\cref{sec:experiments}).


\section{Method: \faider}
\label{sec:method}
\faider{} separates threshold selection from generation control. It has two stages: offline calibration and online decoding.
Offline calibration constructs cross-group references and fixes an empirical
disparity threshold. Online decoding then uses that threshold through either
hard feasibility checks or an adaptive soft penalty during list construction.

\subsection{Offline Calibration with Cross-Group References}
\label{sec:offline}
Let $\mathcal{D}_{\mathrm{cal}}=\{(x_i,a_i)\}_{i=1}^n$ be the calibration
set, disjoint from the test set. For each calibration example, let $y_i$ denote its reference item. We define cross-group neighborhood and semantic disparity as below.

\paragraph{Cross-group neighborhood.} For any $(x,a)$, the set of calibration
examples with similar contexts but a different protected attribute is
\begin{equation}
\Ncal(x,a) = \big\{\, j : \cos(\Emb(x),\Emb(x_j)) \ge \tau_\rho \ \wedge\ a_j\ne a \,\big\}.
\end{equation}

\paragraph{Semantic disparity.} For a candidate output $y$, its disparity relative
to cross-group references is
\begin{equation}
\Delta(x,a;y) = \max_{j\in\Ncal(x,a)} \big\lVert \Emb(y)-\Emb(y_j) \big\rVert_2 .
\label{eq:disparity}
\end{equation}
The maximum measures the largest semantic discrepancy within the selected
cross-group neighborhood. It compares different users with similar contexts,
rather than paired attribute changes for the same context. If the neighborhood
is empty, the score is undefined. Deployment therefore requires a documented
fallback, such as abstention or a broader similarity threshold. Our analysis
assumes nonempty neighborhoods.

\paragraph{Empirical quantile threshold.}
Using the same reference bank, we compute
$\Delta_i=\Delta(x_i,a_i;\hat y_i)$ for the calibration examples and select
their upper empirical quantile at nominal level $1-\alpha$ as $\delta$.
The reference bank and $\delta$ remain fixed during test-time decoding.
Although $a_j\ne a_i$ excludes each example from its own neighborhood,
calibration outputs can still contribute to one another's scores.

The quantile construction is motivated by split conformal calibration
\cite{angelopoulos2023gentle}. Under exchangeability of calibration and new
scores, its corrected order statistic
$\delta=\Delta_{(k)}$, with $k=\lceil(n+1)(1-\alpha)\rceil$ and
$\delta=+\infty$ when $k>n$, yields marginal exceedance probability at most
$\alpha$. In our reference-reuse protocol, however, calibration outputs
help construct the scores being calibrated, whereas a new output does not
enter that reference bank. Calibration/test separation alone therefore does
not establish the score exchangeability required by this argument.
We treat the selected quantile as a \emph{conformal-inspired empirical
cutoff}, without claiming distribution-free finite-sample test coverage for
this construction. Adaptive decoding also changes the output-selection
mechanism. The online bound in \cref{sec:theory} is established separately,
conditional on enforceability, and does not rely on offline coverage.

Note that the item is the semantically correct unit. A recommendation's unit of
meaning is the item (``\textit{Blade Runner}''), not the token. Our disparity $\Delta$ is defined on item-list embeddings precisely because that
is where semantic meaning lives. Controlling at the item level therefore matches the granularity the fairness notion is about. Instead, token-level control would regulate a list-level quantity at a level where it may not even be defined. 
\footnote{%
\label{foot:whitebox}%
When full-vocabulary logits are available (open-weights or self-hosted models), the same penalty can be applied at the token level, $\tilde\ell_t(v)=\ell_t(v)-\lambda\, r_t(v)$, in the style of controlled decoding \cite{Dathathri2020Plug,yang2021fudge}. We do not pursue this. It forfeits black-box compatibility and requires scoring partial generations, whose list-level disparity is ill-defined.}


\subsection{Online Decoding}
\label{sec:family}
Given $\delta$, \faider{} offers two operating points over the same
item-proposal loop:
\begin{itemize}
  \item \textbf{Hard \faider}: return a list only if its final disparity
  satisfies $\Delta\le\delta$; abstain when no feasible beam remains.
  \item \textbf{Soft \faider}: balance likelihood and disparity through an
  adaptive penalty, using completed-list violations to regulate its strength.
\end{itemize}

\subsubsection{Hard \faider: Constrained Item Selection}
\label{sec:hard}

Hard \faider{} maintains a beam of partial recommendation lists.
At each slot, it proposes next items, removes extensions whose disparity
exceeds $\delta$, and retains the top-$B$ feasible extensions by cumulative
log-probability (\cref{alg:hard}). If no extension remains, the decoder
abstains. A final score check ensures that every returned list satisfies the
operational constraint $\Delta\le\delta$. Any optional fallback output must
be identified separately and is not certified by this check.

Prefix filtering is a search heuristic, not a feasibility-complete procedure.
List-level embedding disparity need not be monotone as items are appended,
so a rejected prefix could have led to a feasible complete list.
The guarantee concerns returned lists, not the existence or discovery of a
feasible list for every query, nor the evaluation criterion in \cref{def:mac}.

\begin{algorithm}[t]
\caption{Hard \faider\ (constrained beam search over items)}
\label{alg:hard}
\begin{algorithmic}[1]
\REQUIRE beam size $B$, threshold $\delta$, max items $K$,
proposer \textsc{ProposeItems}, embedder $\Emb$

\STATE $\mathcal{B}_0 \gets \{([\,],0)\}$
\COMMENT{(item list, logprob) pairs}

\FOR{$k=1$ \textbf{to} $K$}
    \STATE $\mathcal{C}\gets\emptyset$
    \FOR{all $(L,S)\in\mathcal{B}_{k-1}$}
        \FOR{all $c\in \textsc{ProposeItems}(x,a,L)$}
            \STATE $L'\gets L\oplus[c]$;
            $\hat\Delta\gets\Delta(x,a;\textsc{Join}(L'))$
            \IF{$\hat\Delta\le\delta$}
                \STATE Add $(L', S+\log p_\theta(c\mid x,a,L))$
                to $\mathcal{C}$
            \ENDIF
        \ENDFOR
    \ENDFOR
    \IF{$\mathcal{C}=\emptyset$}
        \STATE \textbf{return} ABSTAIN (infeasible beam)
    \ENDIF
    \STATE $\mathcal{B}_k\gets$ top-$B$ of
    $\mathcal{C}$ by logprob
\ENDFOR

\STATE \textbf{return} best list in $\mathcal{B}_K$ if final $\Delta\le\delta$; otherwise ABSTAIN

\end{algorithmic}
\end{algorithm}

\subsubsection{Soft \faider\ with Dual-Inspired Feedback}
\label{sec:soft}
Soft \faider{} permits individual threshold violations while targeting their
frequency across queries. It samples $M$ candidate items at each slot and
selects the one with the highest fairness-penalized score. The penalty weight
$\lambda_t$ is held fixed while constructing query $t$'s list and updated
once after that list is complete.

\paragraph{Constrained objective.}
The following population-level objective motivates the controller. For a
decoding policy $\pi$ producing $y\sim\pi(\cdot\mid x,a)$, write
$\Exp_\pi$ for expectation over both queries and policy randomness:
\begin{equation}
\max_{\pi}\ \Exp_\pi\!\big[\log p_\theta(y\mid x,a)\big]
\quad\text{s.t.}\quad
\Exp_\pi\!\big[\one\{\Delta(x,a;y)>\delta\}\big]\le\alpha .
\label{eq:primal}
\end{equation}
This objective specifies the desired trade-off; the implemented decoder is
an approximation, and the proved result concerns empirical violation
frequency rather than an unconditional population-risk guarantee.

\paragraph{Lagrangian and per-slot selection rule.} Introducing a multiplier
$\lambda\ge0$ gives
$\Lcal(\pi,\lambda)=\Exp_\pi[\log p_\theta(y\mid x,a)]-\lambda(\Exp_\pi[\one\{\Delta>\delta\}]-\alpha)$.
We use this formulation to motivate two approximations: replacing the
violation indicator with raw disparity during selection, and making greedy
item-level decisions over a finite proposal pool. For partial list $L$ and
proposals $\{c_m\}_{m=1}^M$, the resulting rule is
\begin{equation}
c^\star=\argmax_{m}\ \Big[\log p_\theta(c_m\mid x,a,L) - \lambda\,\Delta\big(x,a;\textsc{Join}(L\oplus[c_m])\big)\Big].
\label{eq:perslot}
\end{equation}
Here $\oplus$ appends an item, and \textsc{Join} converts the list to its
textual representation. The surrogate provides a graded penalty, but its
minimization is not equivalent to minimizing threshold violations.
Likewise, greedy selection does not solve the full-list Lagrangian
maximization. Fixed-$\lambda$ Soft \faider{} uses \eqref{eq:perslot} with a
constant penalty. The adaptive variant uses the feedback rule below:

\paragraph{Online feedback update.}
After constructing the complete list $y_t$ for query $(x_t,a_t)$, we evaluate
its threshold violation and update the penalty:
\begin{equation}
\lambda_{t+1}=\big[\lambda_t+\eta(\err_t-\alpha)\big]_+,\qquad
\err_t=\one\{\Delta(x_t,a_t;y_t)>\delta\},
\label{eq:dual}
\end{equation}
with step size $\eta>0$. A violation increases $\lambda$ by
$\eta(1-\alpha)$. A non-violation decreases it by $\eta\alpha$, subject to
projection onto the nonnegative half-line. The projection ensures
nonnegativity, not complementary slackness or optimality of the policy.
Because \eqref{eq:perslot} is a surrogate greedy rule, we interpret
\eqref{eq:dual} as \emph{dual-inspired feedback}, rather than exact dual
subgradient optimization.

The update is also motivated by adaptive conformal inference
\cite{gibbs2021adaptive}, but adjusts enforcement strength while keeping
the disparity cutoff fixed. Selection uses raw disparity, whereas feedback
uses the completed list's binary violation. \Cref{alg:soft} summarizes the
procedure. The conditional analysis in \cref{sec:theory} bounds the empirical
frequency of this binary event directly, without requiring an exact
Lagrangian optimizer.

\begin{algorithm}[t]
\caption{Soft \faider\ with dual-inspired feedback (streaming)}
\label{alg:soft}
\begin{algorithmic}[1]
\REQUIRE $p_\theta$, embedder $\Emb$, neighborhoods $\Ncal$,
threshold $\delta$, target rate $\alpha$, step $\eta>0$,
init $\lambda_1\ge0$, items $K$, proposals $M$

\FOR{each incoming query $(x_t,a_t)$, $t=1,2,\dots$}
    \STATE Build $\Ncal(x_t,a_t)$;
    \quad $L\gets[\,]$

    \FOR{$k=1$ \textbf{to} $K$}
        \STATE Sample proposals
        $\{c_m\}_{m=1}^{M}\sim
        p_\theta(\cdot\mid x_t,a_t,\textsc{FormatList}(L))$

        \FOR{$m=1$ \textbf{to} $M$}
            \STATE $r_m\gets
            \Delta(x_t,a_t;\textsc{Join}(L\oplus[c_m]))$
            \hfill \textit{// raw disparity, same units as $\delta$}
        \ENDFOR

        \STATE $c^\star\gets
        \arg\max_m\big[
        \log p_\theta(c_m\mid x_t,a_t,L)
        -\lambda_t\,r_m\big]$;
        \quad $L\gets L\oplus[c^\star]$
    \ENDFOR

    \STATE $y_t\gets\textsc{Join}(L)$;
    \quad $\err_t\gets
    \one\{\Delta(x_t,a_t;y_t)>\delta\}$

    \STATE $\lambda_{t+1}\gets
    \max\!\big(0,\ \lambda_t+\eta(\err_t-\alpha)\big)$
    \hfill \textit{// feedback update, \cref{eq:dual}}

    \STATE \textbf{emit} $y_t$
\ENDFOR

\end{algorithmic}
\end{algorithm}


\paragraph{Implementation conventions.}
We penalize raw disparity on the same scale as $\delta$, avoiding a
step-dependent rescaling of $\lambda$'s effect. Both threshold calibration
and violation feedback use complete lists. Prefix scores guide selection
only. We initialize $\lambda_1$ through a warm-up on the calibration set
before test evaluation.
The reference bank and $\delta$ stay fixed throughout the evaluated stream.

%% file: c7_analysis.tex
\section{Theoretical Analysis}
\label{sec:theory}
We analyze how completed-list feedback regulates the empirical frequency of
disparity violations. The argument separates a deterministic property of the
update from a sufficient condition linking penalty strength to feasible
outputs. It uses the cumulative-error and telescoping perspective of adaptive
conformal inference \cite{gibbs2021adaptive}, but here the updated variable
controls decoding strength while the reference bank and threshold $\delta$
remain fixed.

Throughout, let $\alpha\in(0,1)$, $\eta>0$, and $\lambda_1\ge0$ be fixed, and
let $\err_t=\one\{\Delta(x_t,a_t;y_t)>\delta\}$ be the completed list's
violation indicator. Write
\[
\widehat v_T=\frac{1}{T}\sum_{t=1}^{T}\err_t
\]
for the empirical violation frequency over the first $T$ queries.

\subsection{Feedback and Penalty Growth}

\begin{lemma}[Pathwise feedback bound]
\label{lem:feedback}
For any realized binary violation sequence, the iterates of
\eqref{eq:dual} satisfy, for every $T\ge1$,
\begin{equation}
\widehat v_T
\le \alpha+\frac{\lambda_{T+1}-\lambda_1}{\eta T}.
\label{eq:feedback-bound}
\end{equation}
\end{lemma}
\begin{proof}
Since $[z]_+\ge z$, the update gives
$\lambda_{t+1}\ge\lambda_t+\eta(\err_t-\alpha)$.
Rearranging and summing over $t=1,\ldots,T$ yields
\[
\sum_{t=1}^{T}(\err_t-\alpha)
\le \frac{1}{\eta}\sum_{t=1}^{T}(\lambda_{t+1}-\lambda_t)
=\frac{\lambda_{T+1}-\lambda_1}{\eta}.
\]
Dividing by $T$ proves \eqref{eq:feedback-bound}.
\end{proof}

This inequality requires neither enforceability nor a distributional
assumption on the queries. It identifies penalty growth as the quantity
governing excess violations: $\lambda_{T+1}=o(T)$ suffices for
$\limsup_{T\to\infty}\widehat v_T\le\alpha$, while a uniformly bounded
penalty yields an $O(1/T)$ upper bound on the excess rate.
The next step is to identify a sufficient condition for such boundedness.

\subsection{Conditional Finite-Horizon Control}

\begin{assumption}[Uniform enforceability]
\label{ass:enforce}
There exists a finite $\lambda^\star\ge0$, independent of the horizon $T$,
such that along the entire query and proposal sequence,
\[
\lambda_t\ge\lambda^\star \quad\Longrightarrow\quad \err_t=0
\qquad\text{for every }t\ge1.
\]
\end{assumption}
This condition links penalty strength to final-list feasibility. It is not
ensured by the feedback update alone. The existence of a feasible complete
list is insufficient: a finite proposal pool may omit the required
continuations, and greedy selection using raw disparity need not reach that
list. With randomized proposals, the results below hold pathwise on
realizations satisfying Assumption \ref{ass:enforce}. An almost-sure conclusion would
additionally require this condition to hold almost surely.

\begin{lemma}[Bounded penalty]
\label{lem:bounded}
Under Assumption \ref{ass:enforce}, the iterates satisfy
$0\le\lambda_t\le\lambda_{\max}$ for every $t\ge1$, where
\[
\lambda_{\max}=\max\{\lambda_1,\lambda^\star+\eta\}.
\]
\end{lemma}
\begin{proof}
Projection ensures nonnegativity. For the upper bound, proceed by induction,
starting from $\lambda_1\le\lambda_{\max}$.
If $\lambda_t<\lambda^\star$, then $\err_t-\alpha\le1$ implies
$\lambda_{t+1}\le\lambda_t+\eta<\lambda^\star+\eta\le\lambda_{\max}$.
If $\lambda_t\ge\lambda^\star$, Assumption \ref{ass:enforce} gives $\err_t=0$, and hence
$\lambda_{t+1}=\max\{0,\lambda_t-\eta\alpha\}\le\lambda_t\le\lambda_{\max}$.
Both cases preserve the bound.
\end{proof}

\begin{prop}[Finite-horizon empirical violation control]
\label{prop:violation}
Under Assumption \ref{ass:enforce}, for every $T\ge1$,
\begin{equation}
\widehat v_T
\le \alpha+\frac{\lambda_{\max}-\lambda_1}{\eta T}
\le \alpha+\frac{\lambda_{\max}}{\eta T}.
\label{eq:bound}
\end{equation}
In particular, $\limsup_{T\to\infty}\widehat v_T\le\alpha$.
\end{prop}
\begin{proof}
Apply \cref{lem:feedback} and substitute
$\lambda_{T+1}\le\lambda_{\max}$ from \cref{lem:bounded}.
The second inequality follows from $\lambda_1\ge0$.
\end{proof}

Thus, enforceability supplies the boundedness needed to convert the general
feedback inequality into finite-horizon rate control. Note that the result is one-sided: the violation frequency may remain below $\alpha$, rather than converging to it. Its finite-horizon usefulness depends on the constants. An unknown or large $\lambda^\star$ can make the bound loose.




\paragraph{Remark.}
The results concern empirical exceedances of the decoding surrogate's
threshold, not the paired attribute-change metric in \cref{def:mac},
recommendation utility, or population risk. They require neither offline
conformal coverage nor exact Lagrangian optimization. No exchangeability
assumption is needed, so the pathwise argument permits nonstationary or
adversarial queries. The $O(1/T)$ conclusion nevertheless requires uniform
enforceability throughout the sequence, including after distribution shifts.

%% file: c7_exp.tex
\section{Experiments}
\label{sec:experiments}

\subsection{Setup}
\paragraph{Datasets and evaluation protocol.}
Following \cite{fayyazi2025facter}, we evaluate on MovieLens-1M
\cite{harper2015movielens} and Amazon Movies \& TV
\cite{mcauley2015image}, with gender as the protected attribute.
We split users into disjoint calibration and test sets in a 70/30 ratio.
Held-out items provide the ground truth for recommendation evaluation.
Calibration users supply both the cross-group reference outputs and the
scores used to select $\delta$.
Test users are excluded from reference-bank construction and threshold
selection.

\paragraph{Candidate-set implementation.}
\Cref{alg:soft} presents the sampling-based formulation, in which the LLM
proposes candidate items. The reported experiments use a simplified
candidate-set implementation: recommendations are constructed from supplied
candidates instead of next-item proposals sampled from the LLM.
Fairness-adjusted item selection operates over these supplied candidates.
The results therefore evaluate control within a candidate set, not the
LLM proposal-generation step in \cref{alg:soft}.

\paragraph{Decoding settings.}
We use beam width $B=3$ for Hard \faider{}. Unless otherwise noted, the
neighborhood similarity threshold is $\tau_\rho=0.95$, the feedback step
size is $\eta=0.05$, and the target violation rate is $\alpha=0.05$.
The soft penalty is initialized through a warm-up on the calibration set
before test evaluation. For CFR evaluation, we use decoding temperature
zero when comparing recommendations under protected-attribute changes.

\paragraph{Baselines.}
We compare with three methods. \emph{Zero-Shot} \cite{hou2024large}
uses the LLM without fairness control. \emph{UP5} \cite{hua2024up5}
uses soft prompt tuning to address recommendation fairness.
\emph{FACTER} \cite{fayyazi2025facter} combines conformal thresholding
with iterative prompt repair, providing a comparison with an intervention
that operates through corrective prompting. We evaluate both \faider{}
variants: hard disparity constraints and a soft penalty adjusted through
dual-inspired feedback.

\paragraph{Models and embeddings.}
The recommendation backbones are Qwen3-8B \cite{yang2025qwen3} and
Llama3.1-8B-Instruct \cite{grattafiori2024llama}.
For embedding-based comparisons, we use the SentenceTransformer model
paraphrase-mpnet-base-v2 \cite{reimers2019sentence}, fine-tuned on
recommendation-specific data. Experiments run on NVIDIA A100 GPUs with
80 GB of memory.

\paragraph{Metrics.}
Recommendation quality is measured using Recall@10 and normalized
discounted cumulative gain at rank 10 (NDCG@10). Recall@10 measures the
fraction of held-out relevant items retrieved in the top ten recommendations.
NDCG@10 additionally accounts for their ranking positions.
We measure counterfactual disparity using CFR \cite{fayyazi2025facter},
the mean embedding shift between recommendation lists when a protected
attribute changes while the user context remains fixed. Lower CFR indicates
greater semantic stability under this attribute change.
Higher Recall@10 and NDCG@10 and lower CFR are preferred.

\begin{table}[t]
\centering
\small
\setlength{\tabcolsep}{4pt}
\renewcommand{\arraystretch}{1.1}
\caption{Recommendation quality and counterfactual disparity on MovieLens
and Amazon. \faider{} results use the candidate-set implementation.
The best and second-best reported values for each dataset and backbone
are shown in \textbf{bold} and \underline{underlined}, respectively.}
\label{tab:main}
\resizebox{\textwidth}{!}{%
\begin{tabular}{@{}llccc@{\hspace{10pt}}ccc@{}}
\toprule
\multirow{2}{*}{Dataset}
& \multirow{2}{*}{Method}
& \multicolumn{3}{c}{Qwen3-8B}
& \multicolumn{3}{c}{Llama3.1-8B-Instruct} \\
\cmidrule(lr){3-5}\cmidrule(lr){6-8}
& & Recall@10 $\uparrow$ & NDCG@10 $\uparrow$ & CFR $\downarrow$
  & Recall@10 $\uparrow$ & NDCG@10 $\uparrow$ & CFR $\downarrow$ \\
\midrule
\multirow{5}{*}{MovieLens}
& Zero-shot
& \textbf{0.402} & \textbf{0.458} & 0.742
& \textbf{0.358} & \textbf{0.402} & 0.701 \\
& UP5
& 0.381 & 0.427 & 0.613
& 0.336 & 0.374 & 0.556 \\
& FACTER
& 0.389 & 0.445 & 0.591
& 0.338 & 0.382 & 0.440 \\
& Hard \faider{}
& 0.382 & 0.432 & \textbf{0.389}
& 0.333 & 0.371 & \underline{0.398} \\
& Soft \faider{}
& \underline{0.390} & \underline{0.446} & \underline{0.474} 
& \underline{0.352} & \underline{0.397} & \textbf{0.243} \\
\midrule
\multirow{5}{*}{Amazon}
& Zero-shot
& \underline{0.317} & \textbf{0.351} & 0.814
& \underline{0.296} & \textbf{0.312} & 0.768 \\
& UP5
& 0.294 & 0.328 & 0.721
& 0.261 & 0.289 & 0.672 \\
& FACTER
& 0.301 & 0.339 & 0.634
& 0.277 & 0.279 & 0.581 \\
& Hard \faider{}
& 0.297 & 0.326 & \underline{0.574}
& 0.266 & 0.283 & \underline{0.509} \\
& Soft \faider{}
& \textbf{0.318} & \underline{0.347} & \textbf{0.491}
& \textbf{0.297} & \underline{0.306} & \textbf{0.411} \\
\bottomrule
\end{tabular}}
\end{table}
\subsection{Main Results}

\Cref{tab:main} compares recommendation quality and counterfactual disparity
across the two datasets and two LLM backbones. We have the following observations.

\paragraph{Counterfactual fairness.}
Both variants have lower reported CFR than all three baselines in every
dataset--backbone setting. Soft \faider{} achieves the lowest CFR in three
of the four settings; Hard \faider{} achieves the lowest value on MovieLens
with Qwen3-8B. Relative to FACTER, the soft variant reduces CFR by
$19.8\%$--$44.8\%$. The improvement across both backbones and datasets
supports the usefulness of incorporating a fairness signal into item
selection within the evaluated candidate-set setting.

\paragraph{Recommendation quality.}
Soft \faider{} has higher Recall@10 and NDCG@10 than FACTER in all four
settings and the second-highest NDCG@10 among the compared methods in
each setting. Its NDCG@10 remains close to Zero-Shot, while Recall@10 is
lower on MovieLens and numerically similar on Amazon.
The main benefit is therefore the joint improvement in measured fairness
and ranking quality relative to prompt repair. Sometimes the soft \faider{} even achieves slightly better Recall than the naive baseline. This indicates that the LLM biasedly picks stereotypical items that are not what the user actually liked, and our method can help it.

\paragraph{Comparison between Soft and Hard variants.}
The soft variant has higher Recall@10 and NDCG@10 than the hard variant
throughout, and improves all three reported metrics in three settings.
On MovieLens with Qwen3-8B, hard filtering instead achieves lower CFR
($0.389$ versus $0.474$), at the cost of lower ranking quality.
Thus, stricter enforcement does not consistently minimize the evaluation
metric. This is consistent with the distinction between the reference-based
decoding surrogate in Equation \eqref{eq:disparity} and the paired attribute-change
criterion in \cref{def:mac}: enforcing the former does not directly
optimize the latter.

\subsection{Sensitivity Analysis}
\begin{figure}[t]
    \centering
    \begin{subfigure}[b]{0.48\linewidth}
        \centering
        \includegraphics[width=\linewidth]{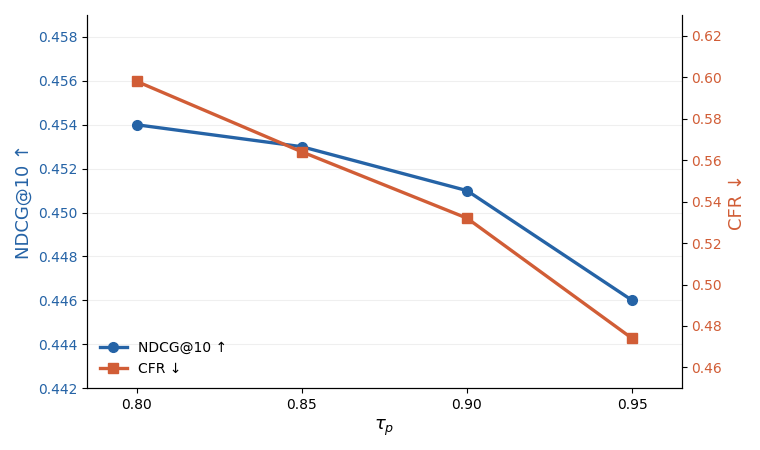}
        \caption{Neighborhood threshold $\tau_\rho$.}
        \label{fig:tau}
    \end{subfigure}
    \hfill
    \begin{subfigure}[b]{0.48\linewidth}
        \centering
        \includegraphics[width=\linewidth]{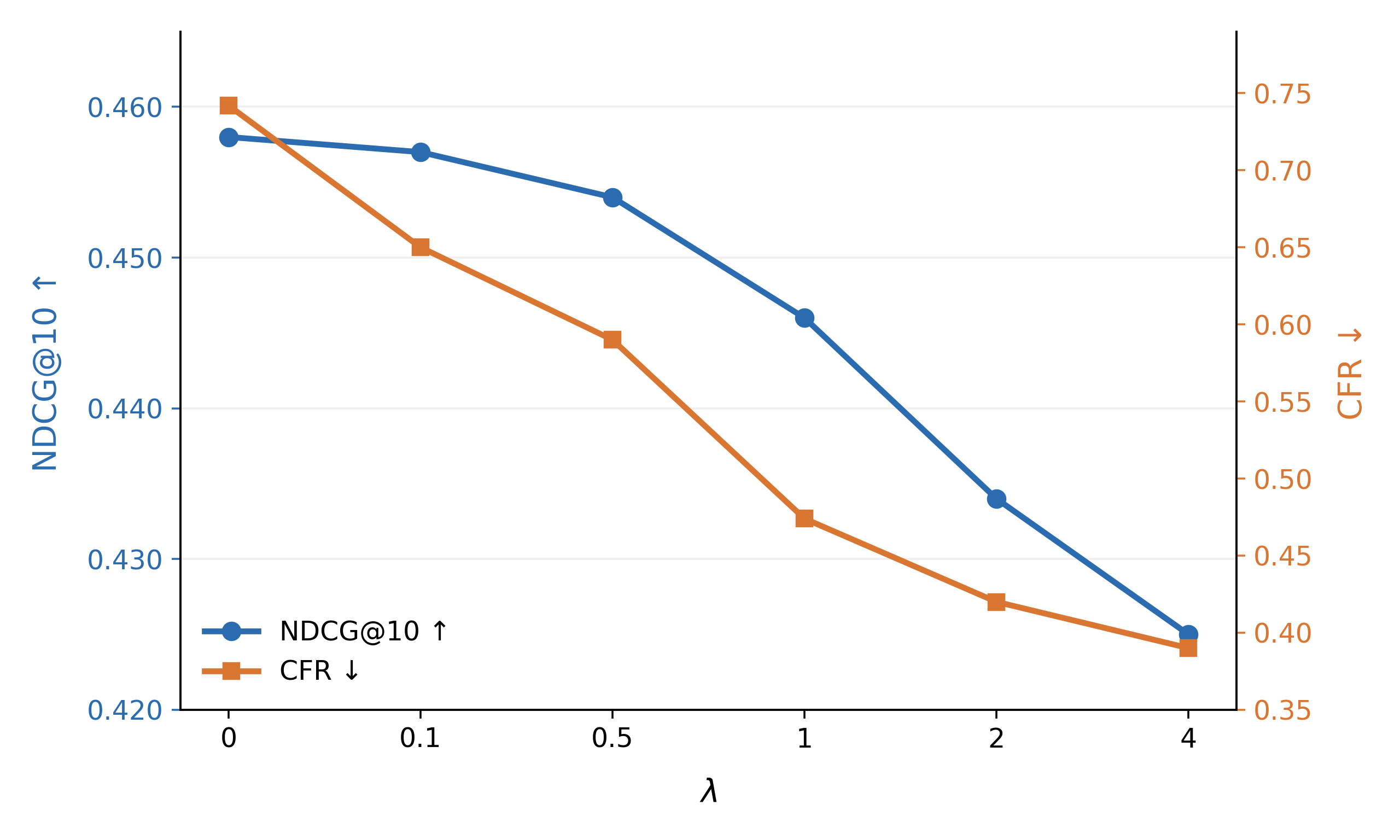}
        \caption{Fixed penalty strength $\lambda$.}
        \label{fig:lambda}
    \end{subfigure}
    \caption{Sensitivity analysis on MovieLens. Each panel reports NDCG@10
    and CFR as the indicated parameter varies; the right panel uses a
    constant penalty rather than online feedback adaptation.}
    \label{fig:hyperparameter}
\end{figure}

\Cref{fig:hyperparameter} examines how neighborhood similarity and fixed
penalty strength affect ranking quality and counterfactual disparity on
MovieLens.

\paragraph{Neighborhood similarity threshold $\tau_\rho$.}
Increasing $\tau_\rho$ restricts the reference neighborhood to cross-group
users with more similar contexts. Across the tested range from $0.80$ to
$0.95$, CFR decreases from $0.598$ to $0.474$, while NDCG@10 decreases from
$0.454$ to $0.446$. Stricter context matching is therefore associated with
lower measured disparity and a modest ranking-quality cost in this
experiment. Unlike the penalty weight, this parameter changes which
cross-group comparisons define the control score. The observed trend does
not imply that narrower neighborhoods are uniformly preferable.


\paragraph{Penalty strength $\lambda$.}
We further evaluate the soft decoder with a fixed penalty strength to isolate the effect of $\lambda$. As shown in the figure, increasing $\lambda$ from 0 to 4 consistently reduces CFR, accompanied by a decrease in NDCG@10. In particular, CFR decreases substantially as $\lambda$ increases, while the reduction in recommendation quality becomes more pronounced at larger penalty strengths. These results demonstrate a clear fairness--utility trade-off: stronger fairness penalties improve counterfactual fairness but may compromise ranking performance. Unlike the fixed-penalty variant, the dual-controlled variant dynamically adjusts $\lambda$ in response to observed fairness violations to target a specified violation rate $\alpha$.

%% file: c7_conclusion.tex
\section{Conclusion}

We presented \faider{}, an item-level framework for fairness-aware LLM
recommendation that separates a fixed empirical disparity threshold from
the strength of decoding-time control. It supports hard filtering and
adaptive soft penalties without retraining the base recommender.
Our analysis connects empirical violation control to penalty growth,
yielding a conditional guarantee under uniform enforceability.
Experiments suggest that soft control
generally offers a more favorable balance between semantic stability and
ranking quality than hard filtering. This finding highlights the distinction
between satisfying a reference-based constraint and achieving the fairness
objective used in evaluation. The broader implication is that decoding-time
fairness should be assessed through its joint effects on semantic stability
and recommendation quality, rather than through constraint satisfaction
alone.